\documentclass[sigconf,manuscript]{acmart}

\usepackage{soul}
\usepackage{url}
\usepackage[utf8]{inputenc}
\usepackage{graphicx, fancybox}
\usepackage{amsmath}
\usepackage{booktabs}
\usepackage{amsmath,amsthm,thmtools,mathtools,amsfonts}
\usepackage{ifthen}
\usepackage{enumitem}

\newtheorem{theorem}{Theorem}

\usepackage{natbib}

\newcommand{\gerr}[1]{\mathrm{Gen}_{\Dist}(#1;S)}

\def\[#1\]{\begin{align}#1\end{align}}
\def\*[#1\]{\begin{align*}#1\end{align*}}

\newcommand{\iHS}{\HS_I}

\newcommand{\lab}{\mathcal{Y}}

\newcommand{\Dist}{\mathcal D}
\newcommand\optparen[1]{\ifthenelse{\equal{#1}{}}{}{(#1)}}
\newcommand{\RiskChar}{R}
\newcommand{\Risk}[2]{\RiskChar_{#1}\optparen{#2}}
\newcommand{\EmpRisk}[2]{\hat \RiskChar_{#1}\optparen{#2}}
\newcommand{\MinRisk}[2]{\RiskChar^{\star}_{#1}\optparen{#2}}

\newcommand{\hERM}{h_{\mathrm{ERM}}}
\newcommand{\hERMi}{\hERM^0}

\newcommand{\bigO}[2][]{\mathcal O #1(#2 #1)}

\newcommand{\Reals}{\mathbb{R}}

\DeclareMathOperator*{\newlim}{\mathrm{lim}\vphantom{\mathrm{infsup}}}
\DeclareMathOperator*{\newmin}{\mathrm{min}\vphantom{\mathrm{infsup}}}
\DeclareMathOperator*{\newmax}{\mathrm{max}\vphantom{\mathrm{infsup}}}

\renewcommand{\lim}{\newlim}
\renewcommand{\min}{\newmin}
\renewcommand{\max}{\newmax}
\renewcommand{\inf}{\newinf}
\renewcommand{\sup}{\newsup}

\renewcommand{\Pr}{\mathbb{P}}
\def\EE{\mathbb{E}}

\newcommand{\defn}[1]{\emph{#1}}

\newcommand{\HS}{\mathcal{H}}

\newcommand{\LHS}{\mathcal{H}_{L}}
\newcommand{\SLHS}{\mathcal{H}_{SL}}
\newcommand{\NNHS}{\mathcal{H}_{NN}}

\newcommand{\ERM}[2][]{\mathrm{ERM}_{#2}\optparen{#1}}

\newcommand{\VCdim}[1]{\mathrm{VCdim}(#1)}

\def\sup{\max}
\def\inf{\min}

\newcommand{\X}{\mathcal X}

\usepackage[hide=true,setmargin=true,marginparwidth=.6in]{marginalia}
\usepackage[capitalize]{cleveref}

\usepackage[most]{tcolorbox}
\newtcbtheorem[crefname={Fact}{Facts}]{Fact}{Fact}{%
  theorem style=plain,
  coltitle=black,
  colback=white,
  colframe=white,
  fonttitle=\bfseries,
  boxsep=0em,
  left=0em,
  right=0em,
  top=0em,
  bottom=0em,
}{fact}

\crefname{lemma}{Lemma}{Lemmas}
\crefname{corollary}{Corollary}{Corollaries}
\crefname{theorem}{Theorem}{Theorems}

\AtBeginDocument{%
  \providecommand\BibTeX{{%
    \normalfont B\kern-0.5em{\scshape i\kern-0.25em b}\kern-0.8em\TeX}}}

\setcopyright{rightsretained}
\copyrightyear{2020}
\acmYear{020}
\acmDOI{}

\acmConference{Preprint}{}{b}
\acmBooktitle{}
\acmPrice{999.00}
\acmISBN{}

\begin{document}

\title[Enforcing Interpretability and its Statistical Impacts: Trade-offs between Accuracy and Interpretability]{Enforcing Interpretability and its Statistical Impacts:\\ 
Trade-offs between Accuracy and Interpretability}

\author{Gintare Karolina Dziugaite}
\affiliation{%
  \institution{Element AI}
}

\author{Shai Ben-David}
\affiliation{%
  \institution{University of Waterloo,}
  \institution{Vector Institute}
}

\author{Daniel M. Roy}
\affiliation{%
  \institution{University of Toronto,}
  \institution{Vector Institute}
}

\renewcommand{\shortauthors}{Dziugaite, Ben-David, and Roy}

\begin{abstract}
To date, there has been no formal study of
the statistical cost of interpretability in machine learning.
As such, the discourse around potential trade-offs is often informal and misconceptions abound.
In this work, we aim to initiate a formal study of these trade-offs.
A seemingly insurmountable roadblock is the lack of any agreed upon definition of interpretability. 
Instead, we propose a shift in perspective.
Rather than attempt to define interpretability,
we propose to model
the \emph{act} of \emph{enforcing} interpretability.
As a starting point, we focus on the setting of empirical risk minimization for binary classification,
and view interpretability as a constraint placed on learning.
That is, we assume we are given a subset of hypothesis that are deemed to be interpretable, 
possibly depending on the data distribution and other aspects of the context.
We then model the act of enforcing interpretability as that of performing empirical risk minimization over the set of interpretable hypotheses.
This model allows us to reason about the statistical implications of enforcing interpretability, using known results in statistical learning theory.
Focusing on accuracy, we perform a case analysis, explaining why one may or may not observe a trade-off between accuracy and interpretability when the restriction to interpretable classifiers does or does not come at the cost of some excess statistical risk.
We close with some worked examples and some open problems,
which we hope will spur further theoretical development around the tradeoffs involved in interpretability.
\end{abstract}

\begin{teaserfigure}
\scalebox{.85}{
  \shadowbox{\includegraphics[height=1.625in]{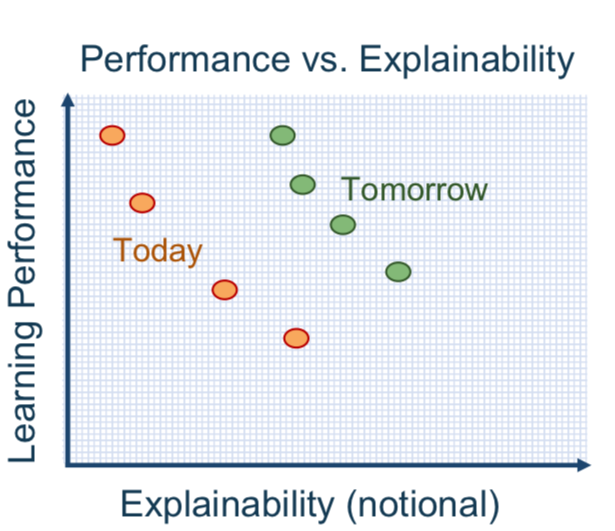}}
  \includegraphics[height=1.775in]{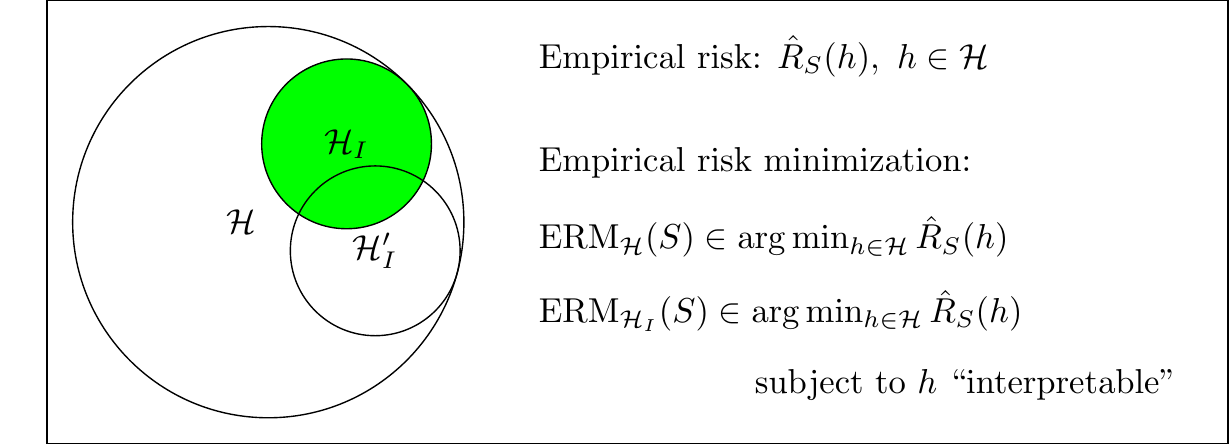}}
  \caption{(left) Figure from DARPA XAI presentation, suggesting inherent tradeoff between explainability and performance. (right) In this work, we model the act of enforcing interpretability as a constraint on learning. We adopt empirical risk minimization as a concrete model for the learning algorithm.}
  \Description{A cartoon plot suggests that intepretability/explainbility comes at the cost of accuracy.}
  \label{fig:teaser}
\end{teaserfigure}

\maketitle

\section{Introduction}

Recent years have witnessed a blowup in the scope of applications of machine learning.
ML-based systems now play a major role in data analysis, prediction/forecasting, and decision-making support.
Among the tasks that machine learning is applied to, many have significant impact on people, especially those involving medical, judicial, and financial decisions.
It is no surprise that, when ML-based algorithms take part in such critical decisions,
there is a demand to understand the way decisions are made---in other words, there is a demand for interpretability.

A naturally arising question is whether there are some inherent trade-offs between the ``interpretability'' of an algorithm and its potential power (be it the scope of situations it can handle, the accuracy of its output or any other measure of performance).
This question is especially pertinent in light of the success of ``black box'' deep learning, 
one of the driving forces behind the adoption of ML.

Although there is no accepted definition of interpretability,
one still finds strong assertions about its relationship with accuracy in the literature.
In particular, 
the idea that, as one demands more interpretability, one suffers in accuracy is 
engrained in the literature on interpretability \citep{caruana2015intelligible,choi2016retain,lakkaraju2016interpretable, tsang2017detecting,wang2019gaining}. At the same time, there is a growing body of work 
suggesting that there is evidence that interpretability does not come at some inherent cost \citep{ribeiro2016should,rudin2018please,heo2018uncertainty}.

Is there only one possible relationship between interpretability and accuracy? What are the mechanisms underlying the relationship between interpretability and accuracy?
Given how important both interpretability and accuracy are to the widespread adoption and success of machine learning and AI, 
it is critical that we approach this question formally.

In this work, we
aim to clarify 
the possible statistical costs of learning interpretable classifiers.
The task of formalizing interpretability, however,
 faces numerous obstacles.

One of the key obstacles is the lack of agreement as to the meaning of interpretability. 
There is a great diversity of approaches to interpretability.
Despite this, we would like to develop theory that is relevant to as broad a swath of work as possible.
One option would be to adopt a popular perspective on interpretability.
For example, much work holds up 
sparse linear predictors as exemplars of interpretability. 
Interpretability is, however, not an inherent property of a classifier, as it depends on numerous factors, including the context, task, and audience \citep{doshi2017towards}.

Rather than attempting to \emph{define} or \emph{quantify} interpretability, 
we instead focus on the \emph{act} of \emph{enforcing} interpretability during learning.
We begin with the framework of empirical risk minimization, 
a simple but powerful approach to learning that is well understood theoretically.
We then view interpretability as placing a \emph{constraint} on learning.
In particular, if $\HS$ denotes the hypothesis set over which we are performing (unconstrained) empirical risk minimization,
we view interpretability as constrained empirical risk minimization over some set $\iHS \subseteq \HS$ 
of ``interpretable'' classifiers.

Rather than working with a particular constraint, associated to a particular notion of interpretability,
we develop a theory that is as agnostic as possible to the constraint.
That is, we aim to draw useful conclusions that depend as little as possible on specific details of $\iHS$.
Indeed, some conclusions that we highlight depend only on the fact that $\iHS$ is a subset of $\HS$.
It is easy to be confused by the level of generality of our arguments. 
Conclusions that hold for arbitrary subsets are consequences of constrained learning and not any specific detail of a notion of interpretability.
It is a mistake, however, to assume that we do not learn anything about interpretability. 
When interpretability functions as a constraint---as it is modeled here---consequences of constrained learning are also those of enforcing interpretability.

The set $\iHS \subseteq \HS$ of interpretable classifiers is presumed to be possibly dependent on the task at hand, the audience interpreting the classifiers, etc. 
We will, however, assume that the determination of whether a classifier is interpretable or not is independent of the training data, $S$. We briefly discuss how this assumption can be relaxed. Note that we do not preclude the use of heldout data in the determination of interpretability. Indeed, any presumed data-distribution dependence
would likely be achieved statistically, using held out data.

Finally, we also touch upon the computational aspects of empirical risk minimization. It is well known
that constrained optimization can be more costly than unconstrained minimization. Viewed from the perspective of interpretability as a constraint, this implies that insisting on interpretability can come at a computational cost as well. We give a concrete example of this, where learning a simple logical formula explaining the data is intractable, while learning a larger, ostensibly less interpretable formula is tractable.

The model we propose here is admittedly simplified, yet we hope that it inspires more sophisticated models and more generally draws attention to the need to build formal models of interpretability and the inherent tradeoffs it brings.

\subsection{Contributions}

\begin{enumerate}
\item We propose to study the \emph{act} of \emph{enforcing} interpretability as a constraint on learning, 
allowing us to work orthogonally to those seeking to define interpretability itself.
\item We seek out conclusions that are agnostic to the particular notion of interpretability being enforced, 
allowing us to derive conclusions that are widely applicable.
\item We provide a summary of key results from statistical learning theory on the framework of empirical risk minimization, linking these to consideration of constrained empirical risk minimization.
\item We provide a case analysis of trade-offs between accuracy and interpretability, showing how each possible relationship can arise.
\item We describe how computational complexity can increase when seeking interpretability through constraints, highlighting another trade-off.

\end{enumerate}

\section{Related work}
\label{sec:relatedwork}

\citet{doshi2017towards} discuss important aspects that need to be considered for quantifying interpretability.
The meaning of a useful or informative interpretation varies with the audience consuming the interpretations, and with the purpose of having the interpretations in the first place. 
The theory developed in this work is valid under any criteria meeting reasonable conditions of interpretability. We discuss this further in \cref{sec:formalize}.

\citet{caruana2015intelligible,choi2016retain} discuss the importance of having interpretable models in healthcare. The authors highlight that uninterpretable models are rejected despite them outperforming interpretable ones, and reiterate that ``a tradeoff must be made between accuracy and intelligibility" \citep{caruana2015intelligible}.
Such statements can be found in multiple other articles outside the healthcare regime (see, e.g., \citep{wang2015trading, wang2019gaining, johansson2011trade, gilpin2018explaining}, among many others).

On the other hand, a number of articles report the existence of models that have no accuracy versus interpretability trade-off. 
\citet{rudin2018please} gives examples of interpretable yet accurate state-of-the-art models, questioning the trade-off and the need for black-box models. The author also gives examples for which choosing interpretable classifiers led to a significant increase in performance.
As another example consider the algorithm proposed in \citep{heo2018uncertainty}, which learns interpretable classifiers that perform just as well as uninterpretable ones on a number of classification tasks in healthcare. \citet{ribeiro2016should} propose a method for explaining classifier's predictions. They find that more interpretable models generalize better.

The most relevant work is by \citet{semenova2019study},
where the authors propose a theoretical tool to identify the existence of simpler-yet-accurate models for a given problem.
This work can be viewed as providing evidence that refutes the interpretability-versus-performance trade-off.
In our case analysis, this setting is captured by the case when the excess approximation error associated to the subclass of interpretable classifiers is small compared to the gains in estimation error.

In \cref{sec:cases}, we discuss how the phenomena discussed in those papers,  observed tradeoffs in some cases and performance gains of interpretabble models in others, is not contradictory and can be clearly understood via the basic principles of our analysis.

\section{Formal Model}
\label{sec:formalize}

We now introduce the formal learning model that we use to study the interpretability--accuracy tradeoff.
In particular, we study supervised classification 
using tools from statistical learning theory.
For a thorough introduction to the concepts discussed here, we refer the readers to the book by \citet{shalev2014understanding}.

In supervised classification, 
we are given training examples in the form of pairs $(x,y)$
where $x$ is an \defn{input} and $y$ is a \defn{label}.
Let $\X$ denote a set of possible inputs 
and let $\lab$ denote a finite set of possible labels.
Let $\X \to \lab$ denote the set of all classifiers, i.e., functions from $\X$ to $\lab$.

For example, 
in a medical image diagnosis problem,
$\X$ could be the set of all bitmap images of a certain dimension,
and $\lab$ might be the two-point set $\{\pm 1\}$,
with $1$ indicating disease is present and $-1$ indicating no disease is present.

On the basis of training examples, we wish to learn a classifier $f \in \X \to \lab$
that will work well on new inputs that we may face in the future.
There are several ways to formalize this. 
In statistical learning theory,
we assume the training examples are randomly chosen. 
In particular, we assume they are independent and identically distributed
from some unknown distribution $\Dist$ on labelled inputs, $\X \times \lab$.
We may then define the probability of misclassification, or \defn{risk}, 
\[
\Risk{\Dist}{h} 
= \Pr_{(x,y) \sim \Dist}[ h(x) \neq y].
\]
While $\Dist$ is presumed to be unknown,
we possess $m$ independent and identically distributed training examples, $S \sim \Dist^m$. 
Writing $S = ((x_1,y_1),\dots,(x_m,y_m))$,
we may define the 
\defn{empirical risk} of any classifier $h \in \HS$ by
\[
\EmpRisk{S}{h} = \frac { \# \{0 \leq i \le m : h(x_i) \neq y_i \} } {m}.
\]
Writing $\EE$ for expectation over samples of a fixed size generated i.i.d. by the distribution $\Dist$,
it is easy to see that, for every sample size $m$ and  every fixed classifier $h$ (i.e., independent from $S$), 
$\Risk{\Dist}{h}=\EE_{S \sim \Dist^m}\EmpRisk{S}{h}$.
Note that this identity does not, in general, hold if $h$ depends on $S$.

\subsection{Interpretability as a Constraint}

In order to study the trade-off between interpretability and accuracy, we must formalize interpretability in some way.
There is, however, no agreed upon definition of interpretability.
Arguably, interpretability can mean very different things in different situations.
This is true even in the context of supervised classification.

As a starting point,
we note that every learning setup implicitly works within some strict subset of the space $\X \to \lab$ of all classifiers. 
We will let $\HS$ denote the space of hypotheses under consideration. It is important to note that $\HS$ will not be assumed to be a subset of $\X \to \lab$. One should think of $\HS$ as a space of \emph{representations} of classifiers. For example, $\HS$ may be the set of all neural network classifiers of some fixed depth and width, represented, e.g, by matrices of real-valued weights. This space is not a function space because two distinct set of weights may represent the same classifier on $\X$. We make this distinction because we will allow for the possibility that one representation of a classifier may be deemed interpretable while another representation may not. 

Rather than attempt to define interpretability, 
we consider instead the effect of \emph{demanding} interpretability, 
i.e., we model interpretability as a constraint on learning.
In particular, we make the simplifying assumption that, for every classifier in $\HS$,
there is a judgment whether the classifier is \emph{interpretable} or \emph{not}.
This approach sidesteps the question of how to define interpretability
and commits only to the idea that this judgment can be made for each classifier. 
This judgment need not be universal---it can be considered to be specific to the problem at hand. 
However, crucially, the judgment is not allowed to depend on the training samples,
although it could depend on some held out samples or the data distribution, if known. That is, the judgmenet whether the classifier is interpretable can be task specific. %

We briefly provide a few examples. If $\HS$ were a set of decision trees, we might define $\iHS$ by a limitation on the size/depth of the tree. If $\HS$ were the set of neural networks of varying depth, then the set of linear predictors $\iHS$ can be obtained as those without a hidden layer. In the context of vision and image classification, we 
might use tools to highlight image regions that most influence the logits of a neural network (say, as measured by certain gradients).
Given a held out data set, we might define $\iHS$ to be the set of classifiers $h$ such that a (theoretical) group of human annotators think that the highlighted regions are sensible. 
Building a formal, analytical description of $\iHS$ in this case would be challenging. Nevertheless, it is a formal subset of $\HS$.

Consider another example based on a frequently encountered approach to interpretability. Under this approach, a classifier $h$ is deemed interpretable if its predictions can be approximated using surrogate classifiers in an ``interpretable'' hypothesis class (such as sparse linear predictors).  We can form our interpretable hypothesis space $\iHS$ as the set of all classifiers $h \in \HS$ that are well-approximated with, say, high-probability, by classifiers in this ``interpretable'' hypothesis class.

This perspective of interpretability-as-constrained-learning clearly does not encompass all approaches to interpretability. 
For example, we do not allow for the possibility that different people may make different judgments about interpretability,
which might motivate one to not model interpretability in a binary way, as we have.
As another example, we do not consider the idea that interpretability should be judged at the level of individual predictions.
Our simplified formalization, however, allows us to make progress on the question of
how working with interpretability \emph{as a constraint} interacts with accuracy.
We return to limitations of our formalization in \cref{sec:discussion}.

Because every classifier in $\HS$ is either interpretable or not,
we may define the subset, $\iHS$, of all interpretable classifiers in $\HS$.
Having formalized interpretability as identifying a subset of a larger hypothesis space,
statistical learning theory immediately yields insight into the possible effects of restricting our attention to interpretable classifiers.
Because these results are agnostic to the particular definition of interpretability, 
we gain insight into the trade-off between interpretability and accuracy for a wide range of situations.

\subsection{Decomposing the error of a learned classifier}

Let $\HS$ be the hypothesis space of classifiers, viewed as functions from a space $\X$ of inputs to a space $\lab$ of labels. 
One can think of this as some set of classifiers chosen to model some observed phenomena, or the set of all
classifiers that some learning algorithm may output (as its input ranges over all possible training samples).

Let $A(S)$ denote the output of the learner on training data $S$.
Note that, because the data $S$ are assumed to be random, 
the hypothesis $A(S)$ is itself a random variable. 
Indeed, the same is true for its risk $\Risk{\Dist}{A(S)}$ and its empirical risk $\EmpRisk{S}{A(S)}$.
As we care about the typical outcome of learning, we are interested in the distribution of these random variables.

Our primary focus is the risk of the learned classifier. 
However, to understand how interpretability relates to risk,
we decompose the risk further.

More formally, let $\MinRisk{\Dist}{\HS} = \inf_{h \in \HS} \Risk{\Dist}{h}$ be the minimum achievable risk by classifiers in $\HS$.\footnote{%
Formally, the minimum may not exist, even if there is a greatest lower bound (i.e., infimum),
and so we should have written $\mathrm{inf}$ rather than $\min$. However, we ignore such issues here. 
Similarly, we use $\max$ in some places where we ought to use $\mathrm{sup}$ to refer to a least upper bound, i.e., supremum.
}
For a learning algorithm $A$, let $\HS_A$ be the class of all classifiers that $A$ may output, given access to any training sample. Then the risk of a classifier $A(S)$ can be decomposed as 
\begin{align*}%
& \Risk{\Dist}{A(S)} \\
&= \MinRisk{\Dist}{\HS_A}
    &&\text{approximation error} \\
&\quad + \Risk{\Dist}{A(S)} - \MinRisk{\Dist}{\HS_A}
    &&\text{estimation error}
\end{align*}

The first component, $\MinRisk{\Dist}{\HS_A}$, is the \defn{approximation error} and is independent of the training sample. 
It is a property of the learning algorithm $A$ (or the class models, or hypotheses, considered).
It may be thought of as the \emph{bias} implied by the choice of learning tool.

The second component, $\Risk{\Dist}{A(S)} - \MinRisk{\Dist}{\HS_A}$, is \defn{estimation error}.
Estimation error quantifies how close we are to the best hypothesis in $\HS$. 
When $h$ is the output $A(S)$ of a learning algorithm, estimation error arises due to overfitting. 
Informally, estimation error arises due to variability in the data (informally, variance).

It is important to highlight that, as a machine-learning user,
we are not interested in the individual errors in risk decomposition in isolation.
Our interest is in the risk of the output classifier. 
However, as we will explain, the interplay between the different types of errors is key to understanding the effects of an interpretability constraint on the risk.

Using the fact that the class of interpretable classifiers is a subset of a larger hypothesis class, i.e., $\iHS \subseteq \HS$,
we can immediately conclude that
\[ 
\MinRisk{\Dist}{\HS} = \inf_{h \in \HS} \Risk{\Dist}{h} \leq \inf_{h \in \iHS} \Risk{\Dist}{h} = \MinRisk{\Dist}{\iHS}.
\]
This leads us to the first fact about the approximation error of interpretable classifiers.
\begin{Fact}{}{approx}
$\HS$ has no larger approximation error than $\iHS$.
\end{Fact}

It follows that the difference between approximation error of $\HS$ and of $\iHS$ quantifies the cost of restricting our attention to interpretable hypotheses, if we ignore estimation error.

The distribution of the estimation error of a learned hypothesis $A(S)$ characterizes the typical gap in risk between $A(S)$ and the best predictor in the class. 
We further analyze the estimation error in the \cref{sec:cases}.

\subsection{Empirical Risk Minimization}

The distribution of estimation error is determined, in part, by the learning algorithm . 
Therefore, in order to study the tradeoffs associated with interpretability, 
we must specify some model for how we intend to learn with and without interpretability as a constraint. 
Arguably, the simplest model to consider is \emph{empirical risk minimization} (ERM) over $\iHS$ and $\HS$.
An algorithm $A$ performs empirical risk minimization over a hypothesis set $\HS$ when,
for all possible data sets $S$,
the learned classifier $A(S)$ achieves the minimum risk, i.e.,
\[
\EmpRisk{S}{\ERM[S]{\HS}} = \inf_{h \in \HS} \EmpRisk{S}{h}.
\]
We refer to a set of classifiers learned by a generic ERM algorithm by $\ERM[S]{\HS}$, with $\hERM$ denoting an element of $\ERM[S]{\HS}$. 
We may then formalize an interpretable learning algorithm as ERM over $\iHS$,
yielding a classifier $\hERMi \in \ERM[S]{\iHS}$.

The \defn{generalization error}, $\gerr{h}$, of a hypothesis $h \in \HS$ is the gap between the risk and empirical risk of $h$, i.e., the difference between the train and test errors. 
Formally,
\[
\gerr{h} = | \Risk{\Dist}{h} - \EmpRisk{S}{h} |.
\]
A key quantity is the \defn{worst-case generalization error over $\HS$}:
\[\label{eq:supgenerror}
 \sup_{h \in \HS} \gerr{h}
\]
It is easy to demonstrate that the estimation error of $\hERM$ is bounded above by twice the worst-case generalization error. To see this, note that, because $\hERM$ achieves the minimal empirical risk, the empirical risk for every $h \in \HS$ is no smaller. Thus, for every $h \in \HS$, this logic justifies the inequality
\begin{align}
&\Risk{\Dist}{\hERM} - \Risk{\Dist}{h} \\
&\quad= \Risk{\Dist}{\hERM} - \EmpRisk{S}{\hERM} + \EmpRisk{S}{\hERM} - \Risk{\Dist}{h} \\
&\quad\leq \gerr{\hERM} + \gerr{h}
\end{align}
Finally, we can bound the sum:
\[
\gerr{\hERM} + \gerr{h} \le 2 \sup_{h' \in \HS} \gerr{h'}.
\]
Since this inequality holds for every $h \in \HS$, 
it holds for the hypothesis $h^* = \arg \min \Risk{\Dist}{h}$ that achieves the minimimum risk. 
Substituting in $h^*$, we obtain the desired bound on the estimation error:
\[
\Risk{\Dist}{\hERM} - \MinRisk{\Dist}{\HS} \leq
2 \sup_{h \in \HS}  \gerr{h} 
\]
\begin{Fact}{}{boundestimation}
The estimation error of $\ERM{\HS}$ is bounded above by twice the worst-case generalization error over $\HS$.
\end{Fact}
Next, we state another fact describing how the generalization error of interpretable classifiers relates to the generalization error of the ones in $\HS$.
Since $\iHS \subseteq \HS$, the following inequality follows trivially:
\[
\sup_{h \in \iHS} \gerr{h} 
  \leq \sup_{h \in \HS} \gerr{h} . 
\]
Therefore, with probability one, the following fact holds:
\begin{Fact}{}{generr}
The worst-case generalization error over $\iHS$ is no larger than the worst-case generalization error over $\HS$.
\end{Fact}
It is worth noting that the generalization error of a particular classifier returned by some learning algorithm may be much better than the worst-case generalization error, even if the algorithm is an ERM.
There is a rich literature on generalization error, and a wide range of tools that have been introduced to study and quantify it
\citep{koltchinskii2000rademacher,bartlett2001rademacher,shawe1998structural, mcallester1999pac,devroye1979distribution,bousquet2002stability,dwork2015preserving,kearns1999algorithmic}.
For binary classification and empirical risk minimization, 
the worst-case generalization error is characterized by the so-called VC dimension \citep{vapnik2015uniform}, 
which we briefly touch on below.

\subsection{Risk decomposition for approximate ERM}
 
In the case of approximate ERM, the output of a learner is determined by three aspects of the learning process:
the search space of hypothesis considered by the learning algorithm, 
the training data (in particular, how representative the data are of the distribution relative to the hypothesis space),
and the computational resources invested.

Accordingly, the error of an output classifier can be decomposed as the sum of the smallest empirical error achievable in the hypothesis space, 
the \emph{optimization error}
(i.e., the gap between the empirical error achieved within the computational resource bounds and the smallest achievable),
and
the \emph{generalization error} (i.e., the gap between the empirical error 
of the output classifier and its true error):
\begin{align*} %
& \Risk{\Dist}{A(S)} \\
&= \min_{h \in \HS} \EmpRisk{S}{h} 
    &&\text{empirical risk of ERM} \\
&\quad + \EmpRisk{S}{A(S)} - \min_{h \in \HS} \EmpRisk{S}{h}
    &&\text{optimization error} \\
&\quad + \Risk{\Dist}{A(S)} - \EmpRisk{S}{A(S)} 
    &&\text{generalization error}
\end{align*}
We will consider this decomposition later when studying the interplay of accuracy (risk) and interpretability.

Like with approximation error, there is a trivial relationship between the ERM risk with and without an interpretability constraint:
\begin{Fact}{}{empriskERM}
The empirical risk of ERM over $\HS$ is no larger than that over $\iHS$.
\end{Fact}

We discuss optimization and generalization error in more detail in the \cref{sec:cases}.

\subsection{Quantifying Expressivity via VC Dimension}
\label{subsec:VC}

The VC dimension is a measure of the expressive power or ``complexity'' of a space of binary classifiers. We start by introducing the notion of \defn{shattering}.

For any subset $X \subseteq \X$ and let $|X|$ denote its size.
Let $\HS \circ X$ denote the set of subsets of $X$ of the form $\{ x \in X : h(x) = 1\}$ for some $h \in \HS$. 
In other words, $\HS \circ X=\{h^{-1}(1)(x) \cap X: h \in \HS\}$ is the set of all possible $0/1$ partitions that classifiers in $\HS$ can induce on $X$.
It follows that $|\HS\circ X| \le 2^{|X|}$. We say that $\HS$ \defn{shatters} a set $X$ if $|\HS \circ X| = 2^{|X|}$.

If all we know about a learning algorithm is that it performs ERM over $\HS$,
then we cannot hope to learn from any data set that $\HS$ shatters.
In these cases, the hypothesis space is ``too complex'' given the number of data.
This logic is formalized by so-called ``No Free Lunch'' theorems.
Thus understanding shattering is critical to understanding the performance of ERM.

The VC dimension of a $\HS$, denoted $\VCdim{\HS}$, 
is the size of the largest shattered set.
The VC dimension tells us the size of the largest training sample  we might obtain perfect classification accuracy, regardless of the true labels.

Assume $\VCdim{\iHS} = d$. 
Let $X$ be a set of $d$ instances that $\iHS$ can shatter. Then since all $h\in\iHS$ are also in $\HS$, $\HS$ can also shatter $S$, and so $\HS$ has a VC dimension of at least $d$.
\begin{Fact}{}{vcinequality}
$\VCdim{\iHS} \le \VCdim{\HS}$.
\end{Fact}

If $\VCdim{\HS} < \infty$, worst-case generalization error over $\HS$ decays to zero as the number of training data grows. This is formalized in the next fact.

\begin{Fact}{}{estimationgap}
With probability at least $1-\delta$ over a training sample $S$ of size $m$,
the worst-case generalization error over $\HS$ satisfies
\[\label{genvcbound}
\sup_{h \in \HS} \gerr{h} \le \bigO[\bigg]{\sqrt{\frac{\VCdim{H}+\log 1/\delta}{m}} }.
\] 
\end{Fact}
It follows immediately from \cref{fact:boundestimation,fact:estimationgap} that the estimation error for ERM satisfies the same inequality in \cref{genvcbound}. We summarize this consequence as follows:

\begin{Fact}{}{}
If $\VCdim{\HS} < \infty$, the risk of every empirical risk minimizer converges to the approximation error of $\HS$ as the number of training data grows.
\end{Fact}

\section{Effect of interpretability on risk}
\label{sec:cases}

In this section, we consider the impact of restricting our attention to interpretable classifiers, using the two risk decompositions presented in Section 3. %
In each case, we describe how a tradeoff between accuracy and interpretability may or may not exist.

\subsection{Approximation and Estimation Error}

We first start by analyzing the decomposition of risk into approximation error and estimation error. %
By \cref{fact:approx}, we know that approximation error can never decrease by restricting one's attention to interpretable classifiers. 
Thus the impact 
of moving from $\HS$ to $\iHS$
on accuracy (risk) is determined by whether the increase in approximation error
\begin{align}\label{appincr}
\MinRisk{\Dist}{\iHS} - \MinRisk{\Dist}{\HS}
\end{align} 
is greater than, less than, or approximately equal to the change in estimation error,
\begin{gather}\label{estdecr}
\begin{split}
\big(\Risk{\Dist}{\ERM[S]{\iHS}} - \MinRisk{\Dist}{\iHS} \big) \quad \\
\quad -
\big(\Risk{\Dist}{\ERM[S]{\HS}} - \MinRisk{\Dist}{\HS}\big).
\end{split}
\end{gather}
As a result, we can find almost any type of tradeoff between accuracy and interpretability.

The two factors determining the estimation error of ERM are the number of training data and the effective capacity of the hypothesis class, the latter being a quantity that, in general, may depend on the data distribution.

In \cref{fact:boundestimation}, we showed that the estimation error of ERM is bounded by twice the worst-case generalization error, 
and in \cref{fact:estimationgap}, we quoted PAC theory bounding the worst-case generalization
error in terms of the ratio of the VC dimension and the number of training data, irrespective of the data distribution.
Thus, if one has a number of training data far in excess of the VC dimension of  $\HS$,
then, irrespective of the data distribution, the estimation error of $\ERM{\HS}$ will be small. In this case, the decrease in estimation error (\cref{estdecr}) 
is small, and so we will see no appreciable tradeoff with accuracy 
(if the increase in approximation error (\cref{appincr}) is small) 
or a tradeoff (otherwise).

When the number of data are not far in excess of the VC dimension of $\HS$, 
then the data distribution comes into play. It may be the case that a class has a large (or infinite) VC dimension, yet, relative to a specific distribution, the capacity (measured, e.g., by the covering number) is small. Indeed, it suffices for the number of training data to be far in excess of this distribution-dependent capacity of $\HS$ for the decrease in estimation error to be negligible. If it is not, then it is possible that, by moving to $\iHS$, 
one could see a significant drop in estimation error. Again, $\iHS$ need only have small capacity for the actual data distribution in question. If the drop in estimation error is large, this could balance or exceed the increase in approximation error, leading to no tradeoff or even a benefit moving to the interpretable class, $\iHS$. In the latter case, we would credit the improved performance on improved generalization.

\subsection{Empirical risk and generalization error}

We can take another view using the decomposition of the risk 
into the empirical risk and the generalization error.
We will focus first on (exact) empirical minimization, deferring 
a discussion of the role of optimization error to later.

By \cref{fact:empriskERM}, we know that the empirical risk of 
$\ERM[S]{\iHS}$ is no smaller than that of $\ERM[S]{\HS}$.
Therefore, like above, 
the impact 
of moving from $\HS$ to $\iHS$
on accuracy (risk) is determined by whether the increase in empirical risk 
\begin{align}
\EmpRisk{S}{\ERM[S]{\iHS}}- \EmpRisk{S}{\ERM[S]{\HS}}
\end{align} 
is greater than, less than, or approximately equal to the sum of the change in generalization error
\begin{gather}
\begin{split}
\big(\Risk{\Dist}{\ERM[S]{\iHS}} - \EmpRisk{S}{\ERM[S]{\iHS}} \big) \quad \\
\quad -
\big(\Risk{\Dist}{\ERM[S]{\HS}} - \EmpRisk{S}{\ERM[S]{\HS}} \big).
\end{split}
\end{gather}

Like with estimation error, the change in generalization error can depend on the data distribution, but this dependence vanishes if the number of samples exceeds the effective (i.e., distribution dependent) capacity of the larger class $\HS$.

The generalization error is bounded by the worst-case generalization error and so the difference in generalization errors can also be bounded in terms of twice the worst-case generalization error of the larger class, $\HS$. Thus, if the number of training examples is great enough, then will be no penalty in terms of generalization error in moving to the larger class, or, in other words, no advantage in terms of generalization error in moving to the interpretable class, $\iHS$. If there is some increase in approximation error, we may expect an increase in empirical risk of ERM over $\iHS$ as compared with $\HS$, and thus a decrease in accuracy overall. 

When the number of data are moderate, however, the interpretable class may provide much smaller generalization error (or it may not, as we address below in our discussion of fast rate bounds). If the generalization error of $\iHS$ is much less, this advantage may make up for the increase in empirical risk (which may be due to increased approximation error). In this case, interpretability has a regularizing effect and leads to improved accuracy.

Heretofore, we have focused on exact ERM, whereas in practice ERM over a hypothesis class can be computational intractable. Instead, one often relies upon an approximate implementation of ERM. In \cref{sec:efficientexample}, we present an example where, even though the approximation error does not change when moving to the interpretable subclass, the optimization error prevents one from learning.

\subsubsection{Fast-rate bounds}

Up to constants, standard generalization bounds converge to zero
at a so-called ``slow'' rate of $O(\sqrt{C/m})$, where $C$ is the ``capacity'' of $\HS$, which can be distribution-dependent. For almost all notions of capacity, we have $C_I \le C$, where $C_I$ is the capacity of $\iHS$. As such, we might naively expect there to be an advantage moving to $\iHS$ when in comes to generalization error. 

However, this discussion ignores the effect of the size of risk, $\Risk{\Dist}{\ERM[S]{\HS}}$, on the generalization error. In particular, when the risk is small, we may be able to obtain ``fast rate'' bound that converges to zero as $O(C/m)$. Thus, if the approximation error of $\HS$ is close to zero, but the approximation error of $\iHS$ is nontrivial, the improvement in capacity may be swamped by the change from a fast to a slow rate of convergence.

\section{Example: moving to a larger class for efficient learning via ERM}
\label{sec:efficientexample}

Interpretability is not just a restriction of the expressive power of hypotheses, it is also a restriction
on the representation of the classifiers. The same classifier may be represented in many ways, some more obscure and others more interpretable. A representation restriction makes it computationally
harder to come up with a good classifier. The following example borrowed from \cite{shalev2014understanding} shows that such a
hardness gap may turn a computational feasible learning task into
a computationally unfeasible one. 

Consider the class of $3$-term disjunctive
normal form formulae (3-DNF); The instance space is $\X = \{0,1\}^n$
and each hypothesis is represented by the Boolean formula of the form
$h(x) = A_1(x) \vee A_2(x) \lor A_3(x)$, where each $A_i(x)$ is a
Boolean conjunction (and AND of Boolean variables or negations of such).  The output
of $h(x)$ is $1$ if either $A_1(x)$ or $A_2(x)$ or $A_3(x)$ output
the label $1$. If all the three conjunctions output the label $0$ then
$h(x)=0$.

Let $H_{3DNF}^n$ be the class of all such 3-DNF formulae.
The size of  $H_{3DNF}^n$ is at most $3^{3d}$. Hence, the sample
complexity of learning  $H_{3DNF}^n$ using the ERM rule is at most $
3d\log(3/\delta)/\epsilon$.

While a $3$-term DNF formula may be considered interpretable, 
from the computational perspective, finding a low-error classifier of this form is hard. 
In particular, Pitt et al. [1988] and Kearns et al.\ [1994] %
showed that unless
RP$=$NP, there is no polynomial time algorithm that solves 
the $3$-term DNF learning problems in which $d_n=n$ by providing an output that 
is a $3$-term DNF formula. 

In contrast, once we relax this requirement on the representation of the output hypothesis,
the learning problem becomes feasible.
The key observation behind such a learner is noting that since $\lor$
distributes over $\land$, each $3$-term
DNF formula can be rewritten as: 
\[
 A_1 \lor A_2 \lor A_3 = \bigwedge_{u
  \in A_1,v \in A_2,w \in A_3} (u \lor v \lor w)
\]

Next, let us define: $\psi : \{0,1\}^n \to \{0,1\}^{(2n)^3}$ such that
for each triplet of literals $u,v,w$ there is a variable in the range
of $\psi$ indicating if $u \lor v \lor w$ is true or false. So, for
each $3$-DNF formula over $\{0,1\}^n$ there is a conjunction over
$\{0,1\}^{(2n)^3}$, with the same truth table. Since we assume that
the data is realizable, we can solve the ERM problem with respect to
the class of conjunctions over $\{0,1\}^{(2n)^3}$.  Furthermore, the
sample complexity of learning the class of conjunctions in the higher
dimensional space is at most $n^3 \log(1/\delta)/\epsilon$. Thus, the
overall runtime of this approach is polynomial in $n$.

The resulting representation of the Boolean function is arguably less interpretable (since the resulting conjunctive formula
may contain many terms, and large conjunctions of such triplets variables may be hard to interpret).

In other words, the interpretability requirement, encapsulated by asking for $3$-term DNF classifiers
turns an otherwise feasibly learnable problem into a computationally infeasible one.

\section{Discussion}
\label{sec:discussion}

We have proposed to study the relationship between accuracy (risk) and interpretability using a simplified model of learning. Namely, by focusing on empirical risk minimization and by modeling interpretability as the act of restricting
one's attention to a \emph{subset} of classifiers that are deemed interpretable by some judgement, we can make a precise analysis of the factors that contribute 
to accuracy (risk) and how they are affected when we shift from performing ERM on $\HS$ versus its interpretable subclass $\iHS$.

One open problem posed by our work is understanding the trade-off between interpretability and accuracy 
when the set $\iHS$ of interpretability hypotheses depends on the training sample. 
Our analysis above relies on the independence of $\iHS$ from $S$, and so any dependence does not permit one to make the conclusions we have made.
Recently, Foster et al.\ [2019] %
have studied data-dependent hypotheses sets using a notion of uniform stability.
It would be interesting to consider how some standard approaches to explainability (which might be viewed as the evidence one uses to
make a judgement as to the interpretability of a classifier) 
might be modified to make them ``stable'' and potentially amenable to an analysis using this frame.

\section*{Acknowledgments}

The authors would like to thank 
Homanga Bharadhwaj,
Konrad Kording,
Catherine Lefebvre,
Alexei Markovits, and
Yuhuai Wu
for feedback on drafts of this work.
DMR and SBD are supported, in part, by NSERC Discovery Grants. 
Additional resources used in preparing this research were provided, in part, by the Province of Ontario, the Government of Canada through CIFAR, and companies sponsoring the Vector Institute.

\bibliographystyle{ACM-Reference-Format}
\bibliography{references}

\end{document}